\documentclass[letterpaper]{article} 
\usepackage{aaai24}  
\usepackage{times}  
\usepackage{helvet}  
\usepackage{courier}  
\usepackage[hyphens]{url}  
\usepackage{graphicx} 
\usepackage{natbib}  
\usepackage{caption} 
\usepackage{algorithm}
\usepackage{algorithmic}
\usepackage[table]{xcolor}
\usepackage{amssymb}
\usepackage{amsmath}
\usepackage{booktabs}
\usepackage{array}
\usepackage{multirow}
\usepackage{subcaption}

\usepackage{newfloat}
\usepackage{listings}
\DeclareCaptionStyle{ruled}{labelfont=normalfont,labelsep=colon,strut=off} 
\floatstyle{ruled}
\newfloat{listing}{tb}{lst}{}
\floatname{listing}{Listing}
\usepackage{marvosym}
\nocopyright
\title{SGFormer: Semantic Graph Transformer for \\ Point Cloud-based 3D Scene Graph Generation}
\author {
    Changsheng Lv\textsuperscript{\rm 1,\rm 2},
    Mengshi Qi\textsuperscript{\rm 1,\rm 2}\thanks{Corresponding author.},
    Xia Li\textsuperscript{\rm 2},
    Zhengyuan Yang\textsuperscript{\rm 3},
    Huadong Ma\textsuperscript{\rm 1,\rm 2}
}
\affiliations {
    \textsuperscript{\rm 1}Beijing Key Laboratory of Intelligent Telecommunications Software and Multimedia\\
    \textsuperscript{\rm 2}Beijing University of Posts and Telecommunications
    \textsuperscript{\rm 3}University of Rochester\\
    \{lvchangsheng, qms, mhd\}@bupt.edu.cn; lixia@bupt.cn; zhengyuan.yang13@gmail.com
}

\usepackage{bibentry}

\begin{document}
\maketitle

\begin{abstract}
In this paper, we propose a novel model called SGFormer, \textit{\underline{S}emantic \underline{G}raph Trans\underline{Former}} for point cloud-based 3D scene graph generation. The task aims to parse a point cloud-based scene into a semantic structural graph, with the core challenge of modeling the complex global structure. Existing methods based on graph convolutional networks (GCNs) suffer from the over-smoothing dilemma and can only propagate information from limited neighboring nodes. In contrast, SGFormer uses Transformer layers as the base building block to allow global information passing, with two types of newly-designed layers tailored for the 3D scene graph generation task. Specifically, we introduce the graph embedding layer to best utilize the global information in graph edges while maintaining comparable computation costs. Furthermore, we propose the semantic injection layer to leverage linguistic knowledge from large-scale language model (\textit{i.e.}, ChatGPT), to enhance objects' visual features. We benchmark our SGFormer on the established 3DSSG dataset and achieve a 40.94\% absolute improvement in relationship prediction's R@50 and an 88.36\% boost on the subset with complex scenes over the state-of-the-art. Our analyses further show SGFormer's superiority in the long-tail and zero-shot scenarios. Our source code is available at https://github.com/Andy20178/SGFormer.
\end{abstract}
\section{Introduction}

\label{sec:intro}
\begin{figure}[t]
    \centering
    \includegraphics[width=1\linewidth]{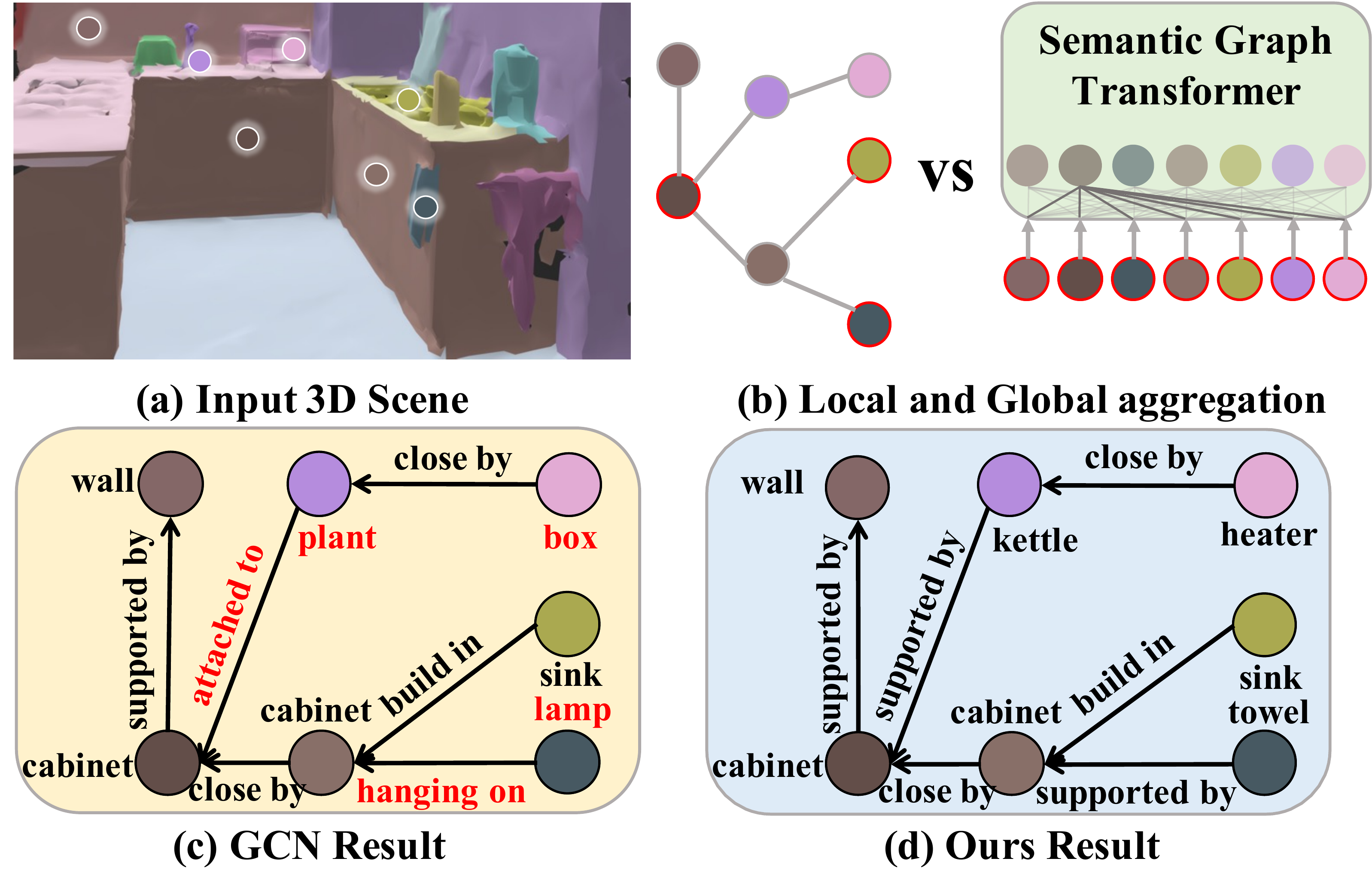}
    \caption{An example of 3D scene graph generation (SGG). (a) an input 3D scene, (b) global aggregation using Transformer on the right \emph{versus} local aggregation using GCN on the left, (c) result using EdgeGCN~\cite{edge-gcn}, and (d) result using our SGFormer. The incorrect results are highlighted in red font, whereas the correct ones are shown in black. SGFormer uses global information aggregation to correct predictions on objects with less information about their neighbors.
    }
    \label{fig: an overview of an example of SGG}
    \vspace{-3mm}
\end{figure}

Understanding a 3D scene is the essence of human vision, requiring accurate recognition of each object's category and localization, as well as the complex intrinsic structural and semantic relationship. 
Conventional 3D scene understanding tasks, such as 3D semantic segmentation~\cite{qi2017pointnet,engelmann2017exploring,rethage2018fully,hou20193d,vu2022softgroup}, object detection and classification~\cite{qi2017pointnet,zhao20193d,zheng2022hyperdet3d}, focus primarily on the single object localization and recognition but miss higher-order object relationship information, making it challenging to deploy such 3D understanding models in practice. To close this gap, the 3D Scene Graph Generation task~\cite{3dssg} is recently proposed, as a visually-grounded graph over the detected object instances with edges depicting their pairwise relationships~\cite{yu2017visual} (as shown in Figure~\ref{fig: an overview of an example of SGG}). The significant importance of 3D scene graph is evident as it has already been applied in a wide range of 3D vision tasks, such as VR/AR~\cite{tahara2020retargetable}, 3D scene synthesis~\cite{dhamo2021graph}, and robot navigation~\cite{gomez2020hybrid}.

One core challenge for 3D scene graph generation is to accurately predict complex 3D scene structures from sparse and noisy point cloud inputs. Most existing methods~\cite{3dssg,edge-gcn,zhang2021knowledge,chen2022exploring} are built based on Graph Convolution Networks (GCNs)~\cite{kipf2016semi}, where nodes represent objects in the scene and edges indicate relationships between objects. Despite the recent success, GCN suffers from its inherent limitation of the over-smoothing dilemma~\cite{li2018deeper}. That is, GCN is effective in modeling neighboring nodes with a controlled number of layers but could struggle with learning the global structure and high-order relationships of the entire scene.
As shown in Figure~\ref{fig: an overview of an example of SGG}, the GCN-based method deteriorates in relationship accuracy when the scene becomes more complicated. 

Increasing the GCN layers or node radius could be one attempt to improve the global modeling capability, but often makes the network difficult to converge and results in worse performance, as discussed in previous studies~\cite{ying2021Transformers} and our later analyses. 
In contrast, we explore a more fundamental change, \textit{i.e.}, replacing the base building block from GCN to Transformer layers~\cite{vaswani2017attention}, which has shown a strong global context modeling capability and therefore could overcome the limitations of GCNs. 

With the above analysis, we propose a Transformer-based model called Semantic Graph TransFormer (SGFormer), to generate a scene graph for a 3D point clouds scene. SGFormer takes both node and edge proposal features obtained from PointNet~\cite{qi2017pointnet} as inputs and stacks Transformer layers to model the higher-order inter-object relationships. The node and edge classifiers then predict each node and edge's categories. However, one technical challenge is the number of edges will grow quadratically to the number of nodes, leading to a quadratically growing input sequence length to the Transformer. To address this challenge, we propose the Graph Embedding Layer that injects the edge information only to the relevant nodes via edge-aware self-attention. In this way, SGFormer preserves the global context with a comparable computation cost.

Furthermore, we propose to exploit external semantic knowledge via the Semantic Injection Layer. Previous GCN-based methods~\cite{zhang2021knowledge,chen2022exploring} have shown the benefit of learning categorical objects and relationships semantic knowledge embeddings. However, these works need to train an extra module to obtain such knowledge. Moreover, the prior knowledge learned from the training set may only work well for common categories in training set distribution but may fail to help with the long-tail object and relationship categories. Alternatively, we design a Semantic Injection Layer to enhance the visual features of object nodes. Specifically, we expand the objects' category labels to detailed descriptions generated from the pre-trained large-scale language model (LLM)~\cite{brown2020language,touvron2023llama}, and extract the text features as semantic knowledge embeddings. Note that our approach requires no extra module and significantly improves SGFormer's performance in long-tail and zero-shot scenarios.

Our main contributions can be summarized as follows:

\par\textbf{(1)} We propose a Semantic Graph Transformer (SGFormer) for 3D scene graph generation, which captures global dependencies between objects and models inter-object relationships with the Graph Embedding Layer. To the best of our knowledge, SGFormer is the first Transformer-based framework for this specific task.

\par\textbf{(2)} We introduce a Semantic Injection Layer to enhance object features with natural language knowledge and show its effectiveness in the long-tail and zero-shot scenarios.

\par\textbf{(3)} We benchmark our SGFormer on the established 3DSSG dataset~\cite{3dssg}, with 40.94\% absolute improvements over the state-of-the-art GCN-based method. More importantly, we achieved an 88.36\% improvement on the subset with complex scenes.


\section{Related Work}

\noindent{\bf Scene Graph Generation.}~Scene graph was first introduced into image retrieval~\cite{johnson2015image} to capture more semantic information about objects and their inter-relationships. Afterward, the first large-scale dataset, Visual Genome~\cite{krishna2017visual}, with scene graph annotations on 2D images gave rise to a line of deep learning-based advances~\cite{xu2017scene,li2017scene,herzig2018mapping,yang2018graph,zellers2018neural,qi2019attentive}, and also contributed to the research of other downstream tasks~\cite{qi2018stagnet,qi2019attentive,qi2019sports,qi2020stc,qi2021latent,qi2021semantics}. 
Nowadays, to understand the complex 3D indoor structure, 3DSSG dataset~\cite{3dssg} was first presented to tackle 3D scene graph~\cite{edge-gcn,zhang2021knowledge} from point clouds. The EdgeGCN~\cite{edge-gcn} introduced an edge-oriented GCN to learn a pair of twinning interactions between nodes and edges, and ~\cite{zhang2021knowledge} divide the generation task into two stages, the prior knowledge learning stage and the scene graph generation with a prior knowledge intervention. 
However, the aforementioned methods fall short in modeling the global-level structure of scenes, and increasing the number of GCN layers easily results in over-smoothing problems. We address the task differently by utilizing the edge-aware self-attention in the Graph Embedding Layer to capture adaptable global representation among nodes.

\noindent{\bf Knowledge Representation.}~There has been growing interest in improving data-driven models with external semantic knowledge in natural language processing~\cite{yang2018graph,hinton2015distilling} and computer vision~\cite{deng2014large,li2017incorporating,qi2019ke,lv2023disentangled}, by incorporating semantic cues, (\textit{e.g.,} language priors of object class names) into visual contents (\textit{e.g.,} object proposals) could significantly improve the generation capability~\cite{pennington2014glove,lu2016visual,speer2017conceptnet,liang2018visual}. Early method~\cite{zellers2018neural} uses statistical co-occurrence as extra knowledge for scene graph generation by introducing a pre-computed bias into the final prediction. 
In this work, we propose a simple yet effective Semantic Injection Layer to enhance visual features with semantic knowledge getting from LLMs-expanded description by cross-attention mechanism. 

\begin{figure*}[!t]
    \centering
    \includegraphics[width=1\linewidth]{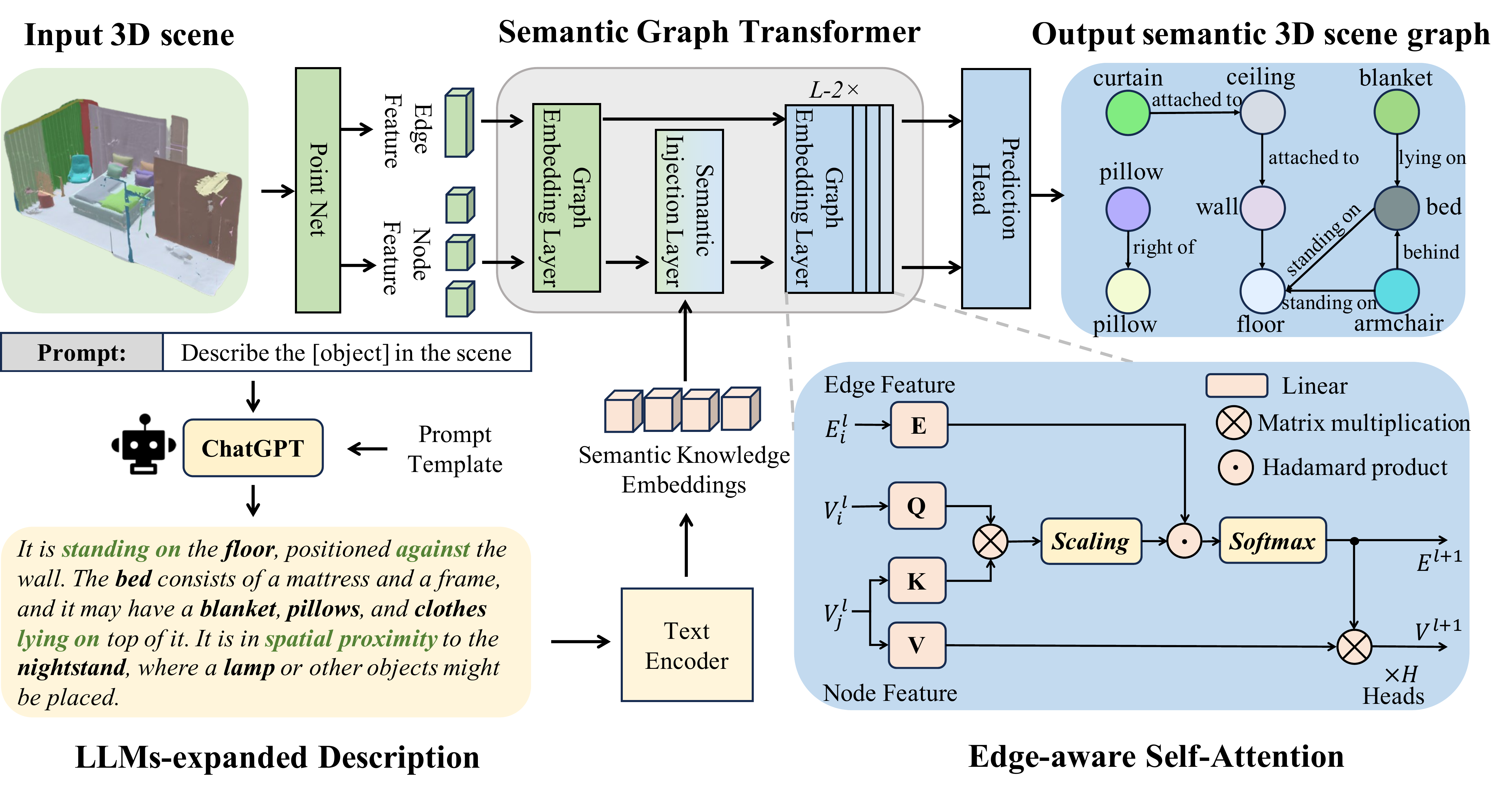}
    \caption{Overview pipeline of our proposed SGFormer. We leverage PointNet to initialize the node and edge features in the 3D scene and use LLMs ( \textit{i.e.}, ChatGPT) to enrich the object description text of the dataset as semantic knowledge. The SGFormer main consists of two carefully designed components: a Graph Embedding Layer and a Semantic Injection Layer. }
    \vspace{-4mm}
    \label{fig: an overview of model}
\end{figure*}

\section{Proposed Approach}
\subsection{Overview}

\noindent\textbf{Problem Definition.}~We define a \emph{3D scene graph} as~$G=(V, E)$, which describes the category of each object and the corresponding semantic inter-object relationships. In the graph, nodes $V$ refer to the object set, while edges $E$ mean inter-object relationships. Meanwhile, we define the output object class labels as $O = \left\{o_1,...,o_N\right\}, o_i\in \mathcal{C}_{\text{node}} $, where $\mathcal{C}_{\text{node}}$ represents all possible object categories, $N$ is the number of nodes. And the set of inter-object relationships can be defined as $R=\left\{r_1,...,r_M\right\}$, $r_j \in \mathcal{C}_{\text{edge}}$, where $\mathcal{C}_{\text{edge}}$ is the set of all pre-defined relationships classes, $M$ denotes the number of valid edges.

As illustrated in Figure~\ref{fig: an overview of model}, the overall framework of our proposed method follows a typical Transformer architecture but consists of two carefully designed components: \emph{Graph Embedding Layer}~\textbf{(GEL)} encodes the global-level features of nodes simultaneously with edge-aware self-attention; \emph{Semantic Injection Layer}~\textbf{(SIL)} extracts the objects' semantic knowledge from the LLMs-expanded descriptions and enhances the node features with knowledge via cross-attention mechanism. For our SGFormer, we empirically set the $L$ layers, \textit{i.e.} a GEL layer and a SIL layer followed by another $L-2$ layers of GEL. We detail the analyses on $L$ and layer setup in ablation studies.

\subsection{Scene Graph Initialization}

Different from 3D instance segmentation~\cite{vu2022softgroup}, we study the higher-order problem of object relationship prediction in the scene.
Note that we use point cloud data with real instance indexes but without category labels. We follow~\cite{edge-gcn} adopt the PointNet~\cite{qi2017pointnet} backbone to capture the point-wise feature~$\mathbf{X}_{P} \in \mathbb{R}^{P \times C_{\text{point}}}$ from the input point cloud~$\mathbf{P} \in \mathbb{R}^{P \times C_{\text{input}}}$ that forms 3D scene~$S$, where~$C_{\text{point}}$ and~$C_{\text{input}}$ 
denote the channel numbers of point clouds and their extracted point-wise feature, respectively, and~$P$ denote the number of sampling points~\footnote{In our experiments, we set $P = 4096$.}.

{\bf Node and edge feature generation.} Following~\cite{edge-gcn}, for the input scene~$S$, we use the average pooling function~\cite{qi2017pointnet} to aggregate the points in~$\mathbf{X}_{P}$ with the same instance index to obtain the corresponding node visual features~$\mathbf{X}_{V} \in \mathbb{R}^{N \times d_{\text{node}}}$, where $N$ indicates the number of instances in scene~$S$, and $d_{\text{node}}$ means node feature dimensions.

Unlike~\cite{3dssg} uses independent PointNet to extract inter-object relationships, we assume that all objects are connected to each other, and thus we can obtain multi-dimensional edge features based on node features. For each $\mathbf{X}_{E_{(i,j)}} \in \mathbb{R}^{d_{\text{edge}}}$, it denotes the feature for the edge $E_{(i,j)}$ that connects two points from subject $V_i$ toward object $V_j$, and the features can be initialized as the following formula using the concatenation scheme introduced in~\cite{wang2019dynamic}:
\begin{equation}\label{eq2}
    \mathbf{X}_{E(i,j)} = (\mathbf{X}_{V_i} \parallel (\mathbf{X}_{V_j} - \mathbf{X}_{V_i})),
\end{equation}
where $\parallel$ denotes the concatenation operation.

\subsection{Graph Embedding Layer}
\label{Graph Encoding M}

Following~\cite{dwivedi2021generalization}, we feed the input node and edge features into the Graph Embedding Layer. The input node features $\mathbf{X}_{V_i} \in \mathbb{R}^{d_{\text{node}}}$ and edge features $\mathbf{X}_{E_{(i,j)}} \in \mathbb{R}^{d_{\text{edge}}}$ are passed via linear projections to embed into $d$-dimensional hidden features $\mathbf{V}_i^0$ and $\mathbf{E}_{(i,j)}^0$, respectively.
\begin{figure}[!t]
    \centering
    \includegraphics[width=1\linewidth]{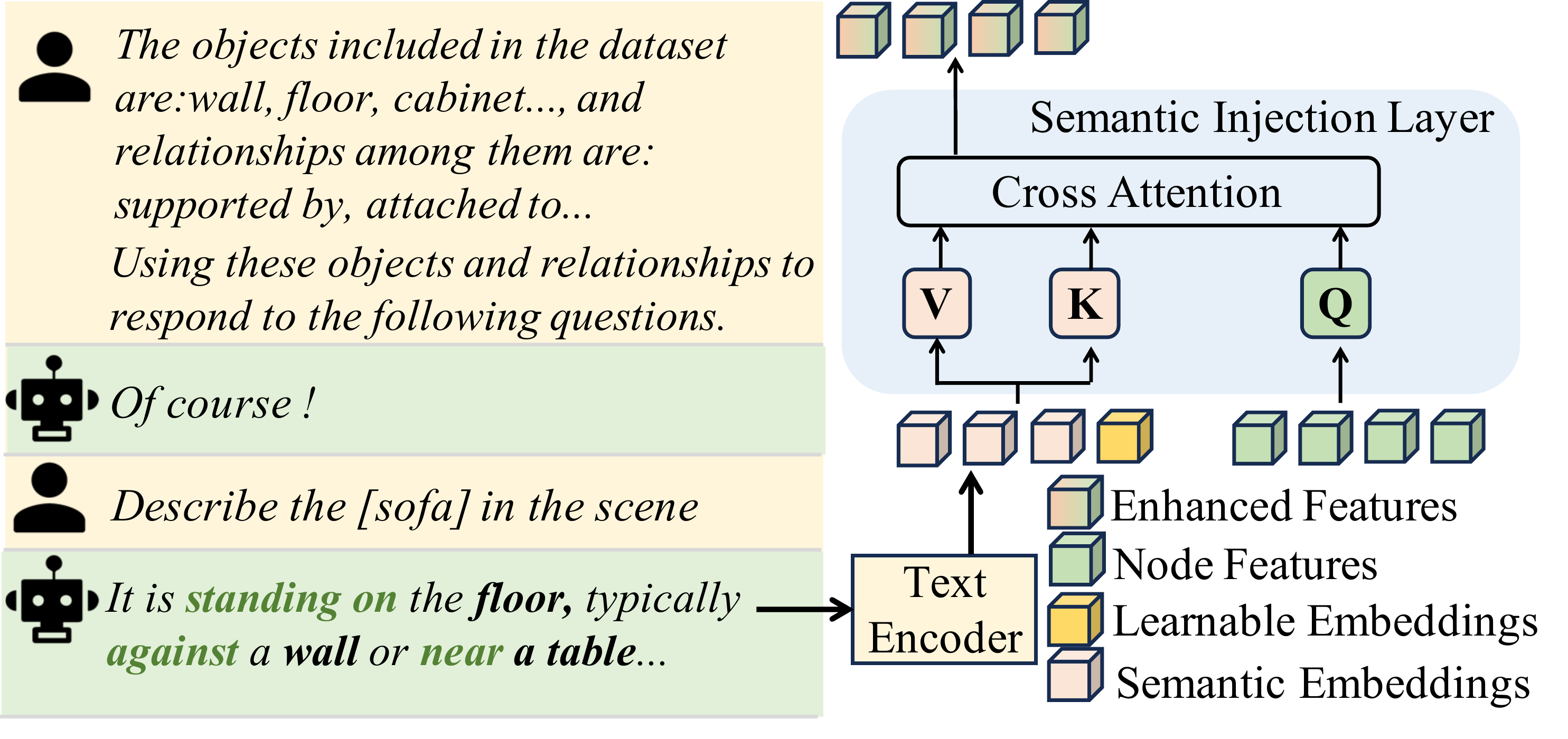}
    \caption{Illustration of the prompt templates to LLMs and the proposed Semantic Injection Layer {\bf (SIL)}. In the layer, node features are employed as queries, while semantic embeddings are utilized as keys and values in the cross-attention. The enhanced features obtained through SIL will serve as the subsequent layer's node features.}
    \label{fig:GT}
    \vspace{-4mm}
\end{figure}

{\bf Multi-Head Edge-aware Self-Attention} is proposed in the layer for the message passing in the graph, which is different from the conventional self-attention described in~\cite{vaswani2017attention}, as shown in Figure~\ref{fig: an overview of model}. For the $l$-th layer, we use the node feature $\mathbf{V}_i^l$ as query and the neighboring node features $\mathbf{V}_j^l$ ($j=1,2,\cdots, N$) as keys and values. 
The node features $\mathbf{V}_i^{l+1}$ in the $l+1$ layer are calculated as the concatenation of the self-attention results from $H$ heads. Besides, the updated edge feature $\mathbf{E}_{(i,j)}^{l+1}$ can be calculated by concatenating the edge-aware self-attention maps ${\mathbf{M}_{ij}^{l,h}} \in \mathbb{R}^{d_h}, 1 \leq h \leq\ H$, $h$ denotes the number of attention heads and $d_h$ denotes the dimension corresponding to each head. Note that, $\mathbf{M}_{ij}^{l,h}$ in our method is a vector instead of a scalar as in the standard Transformer. The information propagating from node $\mathbf{V}_j$ to $\mathbf{V}_i$ can be formulated as the follows in the layer $l$:
\begin{equation}
    \hat{\mathbf{V}}_i^{l+1} =  \mathbf{O}_v^l \cdot \left[ \parallel_{h=1}^H \left( \sum_j^N \mathbf{M}_{ij}^{l,h} \circ \mathbf{W}_V^{l,h} \mathbf{V}_j^{l,h} \right) \right],
\end{equation}
\begin{equation}
    \hat{\mathbf{E}}_{(i,j)}^{l+1} = \mathbf{O}_e^l \cdot \left[\parallel_{h=1}^H \mathbf{M}_{ij}^{l,h} \right],
\end{equation}
where
\begin{equation}
    \mathbf{M}_{ij}^{l,h} = \text{softmax}_j\left(\hat{\mathbf{M}}_{ij}^{l,h}\right),
\end{equation}
\begin{equation}
    \hat{\mathbf{M}}_{ij}^{l,h} = \left(\frac{ (\mathbf{W}_Q^{l,h} \mathbf{V}_i^{l,h})^T \cdot \mathbf{W}_K^{l,h} \mathbf{V}_j^{l,h}}{\sqrt{d_h}}\right) \cdot \mathbf{W}_E^{l,h} \mathbf{E}_{i,j}^{l,h},
\end{equation}
where $\mathbf{W}_Q^{l,h}, \mathbf{W}_K^{l,h}, \mathbf{W}_V^{l,h}, \mathbf{W}_E^{l,h}\in \mathbb{R}^{d_h \times d_h}, \mathbf{O}_v^l, \mathbf{O}_e^l\in \mathbb{R}^{d \times d}, $ $\mathbf{O}_v^l, \mathbf{O}_e^l$ are the weights of linear layers, $\circ$ denotes the Hadamard product, $\cdot$ denotes the matrix multiplication,and $\parallel$ denotes the concatenation operation. 

Following~\cite{dwivedi2021generalization}, in order to keep the numerical stability, the outputs after taking exponents of the terms inside softmax will be clamped to a value in $[-5, 5]$. The outputs $\hat{\mathbf{V}}_i^{l+1}$ and $\hat{\mathbf{E}}_{ij}^{l+1}$ are then separately passed into feed-forward network~(FFN) preceded and succeeded by residual connections and normalization layers. More detail about FFN please refer to~\cite{vaswani2017attention}.

\subsection{Semantic Injection Layer}\label{SIM}
To fully exploit semantic knowledge within the text modality, we design a new prompt template to obtain a detailed description of the object using the LLMs, \textit{i.e.}, ChatGPT~\cite{brown2020language}. After extracting textual features from the descriptions using a text encoder, \textit{i.e.}, CLIP~\cite{radford2021learning}, we fuse them together with visual features by the Cross-Attention mechanism.

{\bf Prompt Template.}~The output of LLMs provides fine-grained descriptions of each object. To align such descriptions with the content of the given dataset, we design prompt templates that include objects and relationships in the dataset. The prompt template is formulated as \textit{``Within the 3D indoor scene dataset, the objects included in it are: [object], and the relationships among them include [relationship]."}, where [object] and [relationship] represent all object and relationship categories in the dataset.
For different objects, we design a prompt as follows: \textit{``Describe the [object] in the scene"}.
For example, we input \textit{``Describe the [floor] in the scene"}, obtaining ChatGPT response as:\textit{``The floor in this 3D scene supporting all the other objects in the room. It lies beneath the bed, chair, sofa, table, and other furniture. The floor is in spatial proximity to the wall, door, and window. It is covered by a rug or carpet, and there might be a towel or clothes lying on it."}

As shown in Figure~\ref{fig:GT}, we obtain descriptions of all objects in the dataset using the aforementioned prompts and extract the semantic embeddings $\mathbf{Z}_K \in \mathbb{R}^{K\times d_{\text{emb}}}$ using a 
frozen pre-trained text encoder, where $K$ is the number of object classes.
And for the unknown object categories that don't have corresponding descriptions, we set learnable embeddings $\mathbf{Z}_U \in \mathbb{R}^{U\times d_{\text{emb}}}$ to represent their text features, where $K + U =|\mathcal{C}_{\text{node}}|$ is the category number and $d_{\text{emb}}$ denotes the text features dimension. The semantic knowledge embeddings set can be formed as follows:

\begin{equation}
    \mathbf{Z} = \mathrm{Concat}(\mathbf{Z}_K, \mathbf{Z}_U).
\end{equation}

\noindent where $\mathrm{Concat}$ refers to concatenation in the column dimension, with the $\mathbf{Z} \in \mathbb{R}^{|\mathcal{C}_{\text{node}}| \times d_{\text{emb}}}$ means the semantic knowledge embeddings of all object categories.

{\bf Cross-Attention} is designed in the layer to aggregate the visual features and the semantic knowledge embeddings followed by two feed-forward networks.
Since the node visual features and the semantic knowledge embeddings are in different latent spaces, we transform node features through a layer of a feed-forward network before fusing them. Then for the node $i$ in $l$-th layer, we use the node feature $\mathbf{V}_i^l$ as query and semantic knowledge embeddings $\mathbf{Z}_i$ as keys and values. We calculate the cross-attention scores $\mathbf{S}_i$ between $\mathbf{V}_i^l$ and each semantic knowledge embedding $\mathbf{Z}_j$ as follows:
\begin{equation}
    \mathbf{H}_i^l = \mathrm{Norm}\left(\mathrm{FFN}_H\left(\mathbf{V}_i^l\right) + \mathbf{V}_i^l\right),
\end{equation}
\begin{equation}
    \mathbf{S}_{i,j}^l = \text{softmax}_j\left(
    \frac{(\mathbf{W}_h \mathbf{H}_i^{l})^T \cdot \mathbf{W}_z\mathbf{Z}_j}{\sqrt{d}}
    \right),
    \label{eq:sim}
\end{equation}
where $\mathbf{W}_h \in \mathbb{R}^{d \times d}$, and~$\mathbf{W}_z \in \mathbb{R}^{d \times d_{\text{emb}}}$. 
Therefore the enhanced node features with semantic knowledge injection can be calculated as follows:
\begin{equation}
    \hat{\mathbf{U}}_i^l = \sum_j \mathbf{S}_{i,j}^l \cdot \mathbf{W}_u \mathbf{Z}_j^{l} ,
\end{equation}
\begin{equation}
    \mathbf{U}_i^l = \mathrm{Norm}\left(\mathrm{FFN}_U\left(\hat{\mathbf{U}}_i^l\right) + \hat{\mathbf{U}}_i^l\right),
\end{equation}
where $\mathbf{W}_u \in \mathbb{R}^{d \times d_{\text{emb}}}$.

\subsection{Training and Inference}

As shown in Figure~\ref{fig: an overview of model}, we first initialize the nodes and edges features and extract embeddings from the LLMs-expanded description of each object to get semantic knowledge embeddings. Then Graph Embedding Layer is used to encode and propagate the global node-edge information, while the Semantic Injection Layer is utilized after the first Graph Embedding Layer to inject the semantic knowledge into the node features. The final result is output from the last Graph Embedding Layer, by the prediction head consisting of two MLPs named NodeMLP and EdgeMLP to recognize objects and their structural relationships, respectively.

Specifically, we adopt the focal loss~\cite{lin2017focal} for objects and predicates classification due to the data distribution imbalance problem~\cite{zhang2021knowledge,3dssg}. We convert object labels into one-hot label and denote prediction logits $p$ as the model's estimated probability for the class where the label $y=1$. Here we define $p_t$:
\begin{equation}
  p_t =
\begin{cases}
p, & \text{if}~y=1 \\
1-p, & \text{otherwise.}
\end{cases}
\label{eq11}
\end{equation}

Hence we formulate the object classification focal loss as the following:
\begin{equation}
    \mathcal{L}_{\text{focal}}^{\text{obj}} = \alpha\left(1-p_t\right)^\gamma \log\left( p_t \right),
\end{equation}
where $\alpha$ is the normalized inverse frequency of objects, and $\gamma$ is a hyper-parameter. Similarly, we can formulate $\mathcal{L}_{\text{focal}}^{\text{edge}}$ as predicate classification focal loss.

Furthermore, to align the injected semantic knowledge embeddings closely with the corresponding object's visual features, we employ the same ground truth in object classification and denote the score $S$ in Eq.~(\ref{eq:sim}) as the estimated probability. Upon deriving $S_t$ from Eq.~(\ref{eq11}), we can define the semantic similarity focal loss for training the Semantic Injection Layer as the following:
\begin{equation}
    \mathcal{L}_\text{SIL} =  \alpha\left(1-\mathbf{S}_t\right)^\gamma log\left( \mathbf{S}_t \right).
\end{equation}

Therefore, the final loss $\textrm{L}_{\text{SG}}$ can be calculated as the sum of the aforementioned three loss functions:
\begin{equation}
    \mathcal{L}_{\text{SG}} = \mathcal{L}_{\text{focal}}^{\text{obj}} + \mathcal{L}_{\text{focal}}^{\text{edge}} + \mathcal{L}_{\text{SIL}}.
\end{equation}

\begin{table*}[!t]
\centering
\footnotesize

\begin{tabular}{lcccccc}
\toprule
&
  \multicolumn{2}{c}{Object Class Prediction} &
  \multicolumn{2}{c}{Predicate Class Prediction} &
  \multicolumn{2}{c}{Relationship Prediction} \\ \cmidrule(r){2-3}  \cmidrule(r){4-5} \cmidrule(r){6-7}
                          \multirow{-2.5}{*}{Graph Reasoning Approach}       & R@5    & R@10   & F1@3  & F1@5  & R@50   & R@100 \\   \cmidrule{1-7}
$+$ $\text{SGPN}$~\cite{3dssg}                           & 89.61  & 96.98  & 63.38 & 77.79 & 32.45  & 41.65 \\
$+$ $\text{GloRe}_{\text{PC}}$~\cite{ma2020global}                       & 84.06  & 95.17  & 69.23 & 80.01 & 31.87  & 42.21 \\
$+$ $\text{GloRe}_{\text{SG}}$~\cite{chen2019graph}                       & 85.27  & 96.62  & 72.57 & 83.42 & 29.58  & 38.64 \\
$+$ $\text{EdgeGCN}^*$~\cite{edge-gcn}                 & 64.46   & 84.88  & 33.34 & 35.90 & 12.19  & 19.69 \\ 
$+$ $\text{EdgeGCN}$~\cite{edge-gcn}                 & 90.70   & 97.58  & \underline{78.88} & \underline{90.86} & 39.91  & 48.68 \\ \cmidrule{1-7}
$+$ $\textbf{Ours w/o SIL}$ & \underline{92.71} & \underline{97.67} & 76.64   & 77.82   & \underline{50.67} & \underline{57.50}   \\
$+$ $\textbf{Ours Full}$  & \textbf{96.41} & \textbf{98.48} & \textbf{85.12}   & {\bf 91.58}   & \textbf{56.25} & \textbf{60.67}   \\ \bottomrule
\end{tabular}

\caption{Comparisons of our model and existing state-of-the-art methods on 3DSSG~\cite{3dssg}. $\text{EdgeGCN}^*$ denotes aggregates information with all nodes, while~$\text{EdgeGCN}$ aggregates information only with nodes that have a known relationship. {\bf Ours w/o SIL} and {\bf Ours Full} denote our model without the Semantic Injection Layer and our full model, respectively. The best performances are shown in bold.
}
\label{table1}
\end{table*}

\section{Experiments}

\subsection{Experimental Settings}

\noindent\textbf{3DSSG Dataset~\cite{3dssg}.} 3DSSG provides annotated 3D semantic scene graphs for 3RScan~\cite{wald2019rio}. 
We follow their RIO27 annotation to evaluate $27$ class objects and $16$ class relationships in our experiments. For the fair comparison, we follow the same experimental settings in~\cite{edge-gcn} and split the dataset into $1084/113/113$ scenes as train/validation/test sets, respectively.

\noindent\textbf{Metrics.} We evaluate our model in terms of object, predicate, and relationship classification. We adopt \emph{Recall@K}~\cite{lu2016visual} for object classification, computing the \emph{macro-F1 score} and \emph{Recall@K} for predicate classification due to the imbalanced distribution. Besides, we multiply the classification scores of each subject, predicate, and object, and then use~\emph{Recall@K} to evaluate the obtained ordered list of relationship classification. Regarding the long-tailed and zero-shot tasks, we use \emph{mean Recall@K(mR@K)}~\cite{tang2020unbiased} for object and predicate classification, and \emph{Zero-Shot Recall@K}~\cite{lu2016visual} for never been observed relationships evaluation.

\noindent\textbf{Compared Methods.} We compare our approach with the following methods on 3DSSG benchmarks: only using PointNet~\cite{qi2017pointnet}, SGPN~\cite{3dssg}, $\text{GloRe}_\text{PC}$ with the point cloud~\cite{ma2020global}, $\text{GloRe}_\text{SG}$ with the scene graph~\cite{chen2019graph}, and Edge-GCN~\cite{edge-gcn}~\footnote{In all experiments, the settings of the above-mentioned methods are adopted from the corresponding papers.}.

\subsection{Implementation Details}
We implement our model based on Pytorch~\cite{paszke2019pytorch} on a single NVIDIA RTX 3090 GPU. Similar to prior works in 3D scene graph generation~\cite{3dssg,edge-gcn}, we choose PointNet~\cite{qi2017pointnet} as the backbone. 
In the Semantic Injection Layer, we use the aforementioned command template to obtain a description of each object, with the category ``objects'' employing a learnable embedding. For the text encoder, we use CLIP and set the $d_{\text{emb}} = 512$.
We set the default layer number $L$ to be $12$, consisting of a pair of GEL and SIL layers followed by $L-2=10$ extra GEL layers. 

\begin{table}[!t]
    \centering
    \begin{tabular}{ccccc}
        \toprule
        Model & Edge                 & Layer & Node R@1                  & Edge R@1                  \\ \cmidrule{1-5}
        Ours full A                     &$\times$              & 3     &68.34                      &88.54                           \\
        Ours full B                     & \checkmark           & 3     &71.07                     & 91.58                    \\
        Ours full C                     & $\times$             & 6     & 69.78                          &89.44                           \\
        Ours full D                     & \checkmark           & 6     & 70.67                  &90.89                      \\
        Ours full E                     & $\times$             & 9     & 71.20                          & 89.85                         \\
        Ours full F                     & \checkmark           & 9     & 72.50                    & 91.86                     \\
        Ours full G                     & $\times$             & 12    & 72.53                          & 93.67      \\
        Ours full H                     & \checkmark           & 12    & \bf{75.33}                     & \bf{96.59}\\ \bottomrule
        \end{tabular}
        \caption{Ablation study on edge feature and layer numbers. 
        }
        \vspace{-6mm}
        \label{GEM}
\end{table}    
\begin{figure*}[!t]
    \centering
    \includegraphics[width = 1\linewidth]{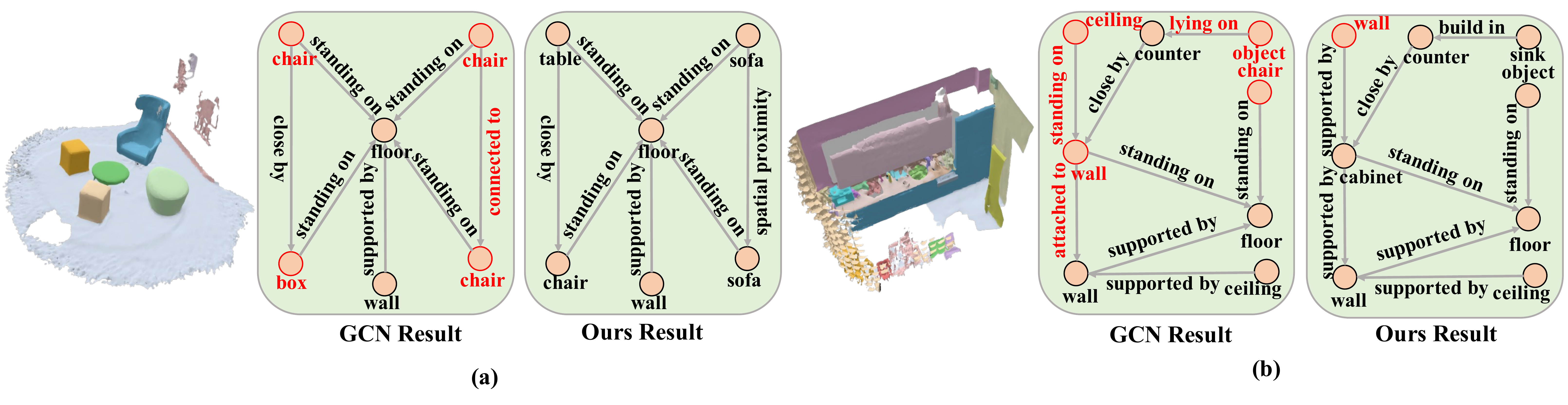}
    \vspace{-8mm}
    \caption{The qualitative results of our model. Given a 3D scene with class-agnostic instance segmentation labels, our SGFormer infers a semantic graph $G$ from the point cloud. For visualization purposes, misclassified object or relationship predictions are indicated in red, while the correct ones are shown in black with the GT value omitted.}
    \label{exp of scene}
    \vspace{-2mm}
\end{figure*} 
\subsection{Results}

\noindent\textbf{Quantitative Results.} We report the quantitative performance of our proposed model compared with other existing methods in Table~\ref{table1}. For \emph{PointNet alone}, we only use PointNet without any reasoning model for evaluation. SPGN~\cite{3dssg} adopts the edges between nodes as a kind of node to conduct message propagation with GCN. As shown in Table~\ref{table1}, using GCN for scene graph generation could improve the object classification but harm the predicate or relationship classification, which confirms the empirical findings reported in~\cite{3dssg,zhang2021knowledge} about the over-smoothing issue caused by multi-layer GCNs. Instead, our model captures the global scene structure with the Transformer architecture, alleviating the over-smoothing in multi-layer GCN and surpassing the state-of-the-art Edge-GCN~\cite{edge-gcn}. Furthermore, compared with the global-level model $\text{GloRe}_\text{SG}$, our proposed model obtains better results by utilizing the information passing between inter-object relationships. Attributing to the introduced Semantic Injection Layer, our model achieves the best result in terms of all evaluation metrics. 

\noindent\textbf{Qualitative Results.}~Figures~\ref{exp of scene} (a) and (b) depict two visualized qualitative results of the state-of-the-art method and our proposed model, and our approach showcases noteworthy advancements in both node and edge prediction. For example, Figure~\ref{exp of scene} (b) depicts a situation when ``sink'' and ``object'' have only one neighbor, which makes EdgeGCN's predictive performance fall short while our proposed SGFormer leverages global information to overcome this challenge and accurately predicting the labels of these nodes. Moreover, EdgeGCN exhibits a cascading effect of erroneous predictions after it hits an error ``ceiling'', while SGFormer delivers more robust results despite encountering the similar situation. These findings highlight the significance of global information aggregation in SGFormer for enhancing the recognition of objects and relationships in complex scenes. Additionally, when two objects have a similar appearance such as ``cabinet'' and ``wall'', EdgeGCN fails to classify them solely based on visual cues. However, our approach is adept at avoiding such errors by incorporating semantic knowledge, indicating the efficacy of our Semantic Injection Layer to provide much-needed semantic assistance.

\subsection{Ablation Studies}
\label{Ablation Studies}

\noindent\textbf{Graph Embedding Layer~(GEL).}
In Table~\ref{GEM}, we design variants of our models as A to H, to examine whether or not to use edge features in the node updates, and the varying layer numbers. Specifically, by comparing the performance of G and H, which have the same depth but differ in whether they use edge features, we can observe that incorporating edge features during graph encoding can significantly enhance the model's predictive capabilities. Additionally, we find that increasing the layer number effectively improves performance, as evidenced by comparing B, D, F, and H, demonstrating the proposed GEL can avoid the over-smoothing problem commonly encountered by GCN when attempting to stack deeper.

\begin{table}[]
    \centering
    \begin{tabular}{lcccc}
    \toprule
    \multicolumn{1}{c}{}                        & \multicolumn{2}{c}{Node}     & \multicolumn{2}{c}{Edge}              \\ \cmidrule(r){2-3} \cmidrule(r){4-5}
\multicolumn{1}{c}{\multirow{-2.5}{*}{Model}} & {R@1} & {mR@5}  &  R@1 & mR@3\\
        \cmidrule{1-5}
        $\text{w/o SIL}$     & 68.80    &84.16  &91.07 &  53.77     \\ \cmidrule{1-5}
        {$ \text{SIL}_{\text{Text}}$}     & 65.51    &83.67  &  87.21  & 53.75      \\
        {$ \text{SIL}_{\text{Basic prompt}}$}     & 71.62   &88.59    &  91.70 & 58.50     \\
        {$ \text{SIL}_{\text{GPT w/o template}}$}     & 73.56   &90.50     & 93.47  & 60.33    \\
        {$ \text{SIL}_{\text{GPT w/ template}}$}     & {\bf 75.33} & {\bf 92.75}     &  {\bf 96.59}  &  {\bf 65.67}   \\ \bottomrule
        \end{tabular}
        \caption{Ablation study for different semantic knowledge. 
        }
        \label{table:SIM}
        \vspace{-4mm}
\end{table}
\begin{figure}[!t]
    \centering
    \includegraphics[width = 1\linewidth]{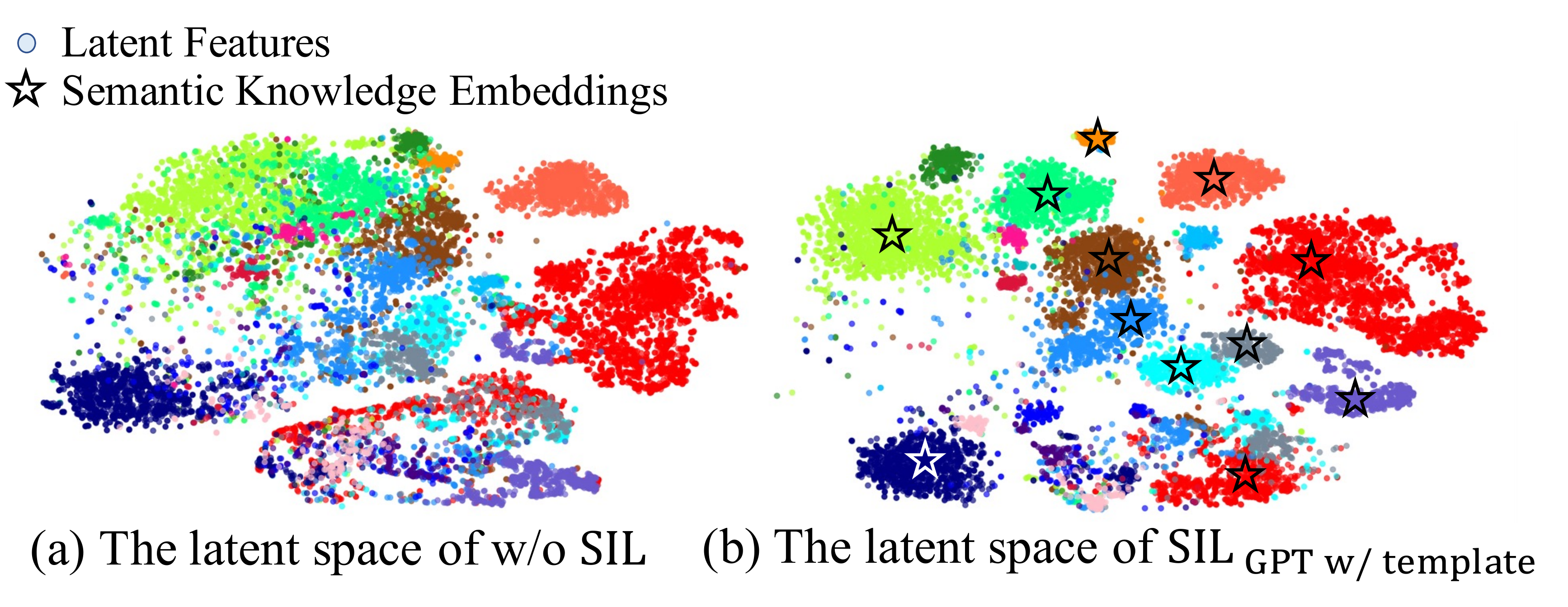}
    \vspace{-6mm}
    \caption{The t-SNE visualization of objects latent space of the w/o SIL(a) and {$ \text{SIL}_{\text{GPT w/ template}}$}(b).
    }
    \label{t-SNE}
    \vspace{-4mm}
\end{figure} 
\noindent\textbf{Semantic Injection Layer~(SIL).} 
We explore various ways to acquire semantic knowledge. To be specific, we directly used object category labels, utilized the basic prompt \textit{``A photo of [object]"}, and obtained descriptions without adding the designed prompt template, denoted as $ \text{SIL}_{\text{Text}}$, $ \text{SIL}_{\text{Prompt}}$, and $ \text{SIL}_{\text{GPT w/o template}}$, respectively. The results of our experiments are shown in Table \ref{table:SIM}. Notably, directly extracting word embeddings from category labels resulted in inferior performance compared to SGFormer without SIL and SIL with the basic prompt. Furthermore, we also find that the injected knowledge from the descriptions with adding the proposed prompt templates can help our proposed model achieve the best performance than a simple prompt and descriptions without adding the prompt templates from Table~\ref{table:SIM}. Besides, Figure~\ref{t-SNE} demonstrates that the output latent features through semantic injection, and we can see our proposed model with SIL using LLMs-expanded description~($ \text{SIL}_{\text{GPT w/ template}}$) achieves a more distinguishable latent space compared with the SGFormer without SIL ($\text{w/o SIL}$), demonstrating that the proposed SIL can enlarge the distance between different object categories, thereby improving object classification accuracy.

\begin{figure}[!t]
    \centering
    \includegraphics[width = 1\linewidth]{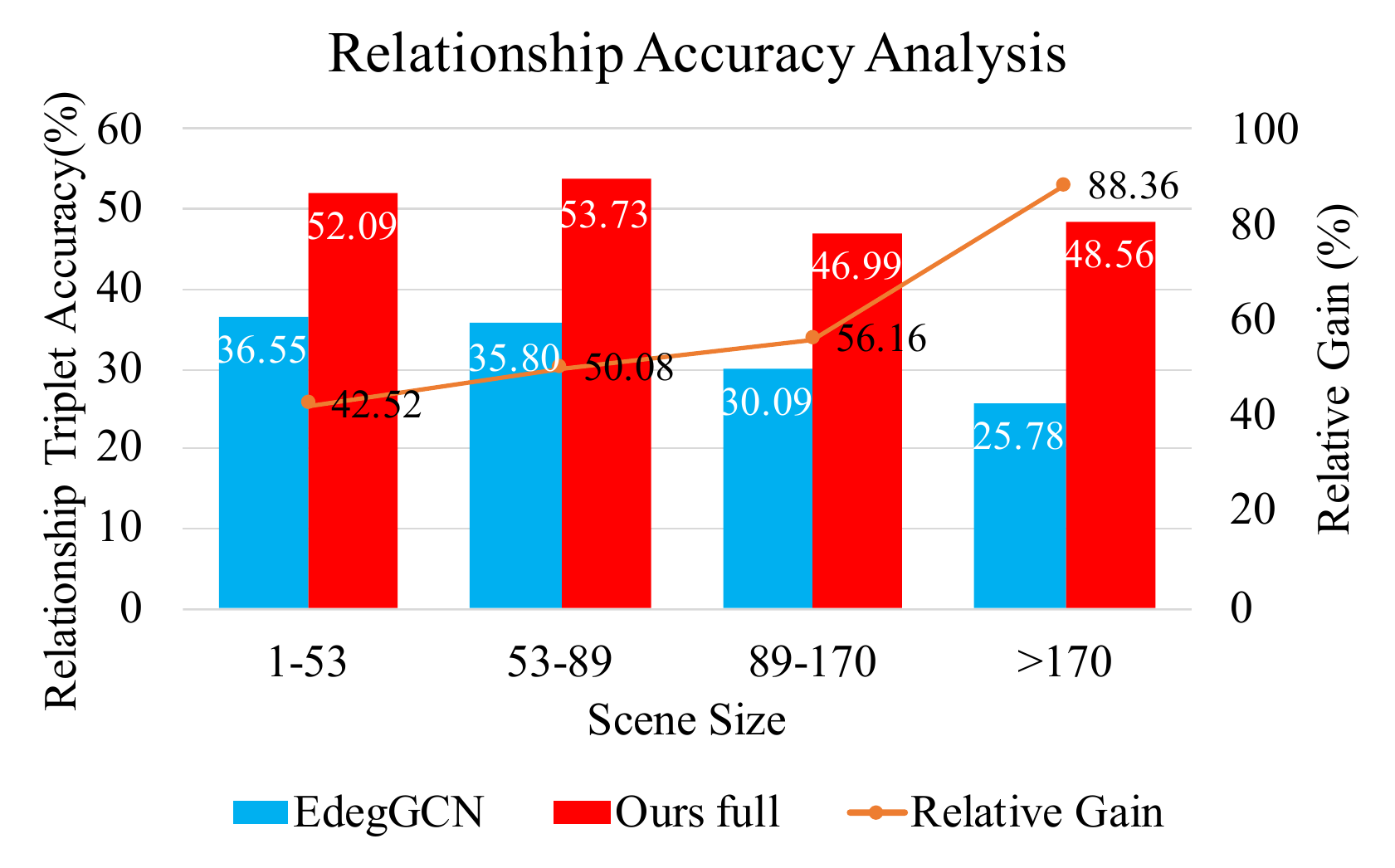}
    \vspace{-6mm}
    \caption{Performance comparison of our full model and EdgeGCN w.r.t relationship classification.}
    \label{Relationship_acc_analysis}
    \vspace{-4mm}
\end{figure}
\begin{figure}[t]
    \centering
    \includegraphics[width = 1\linewidth]{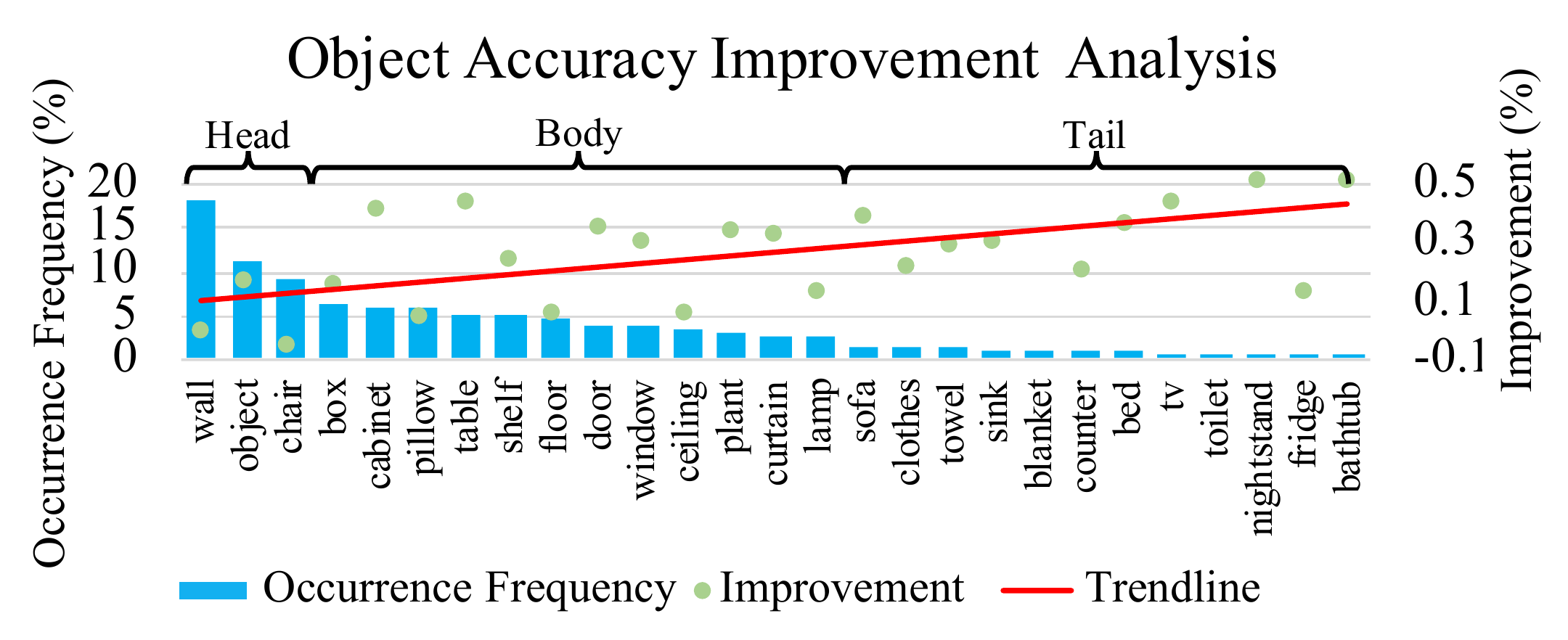}
    \vspace{-6mm}
    \caption{Illustration of objects' per-label accuracy improvement with occurrence frequency in the training set.}
    \label{fig: long-tail-analysis}
    \vspace{-4mm}
\end{figure} 

\begin{figure}[t]
    \centering
    \includegraphics[width = 1\linewidth]{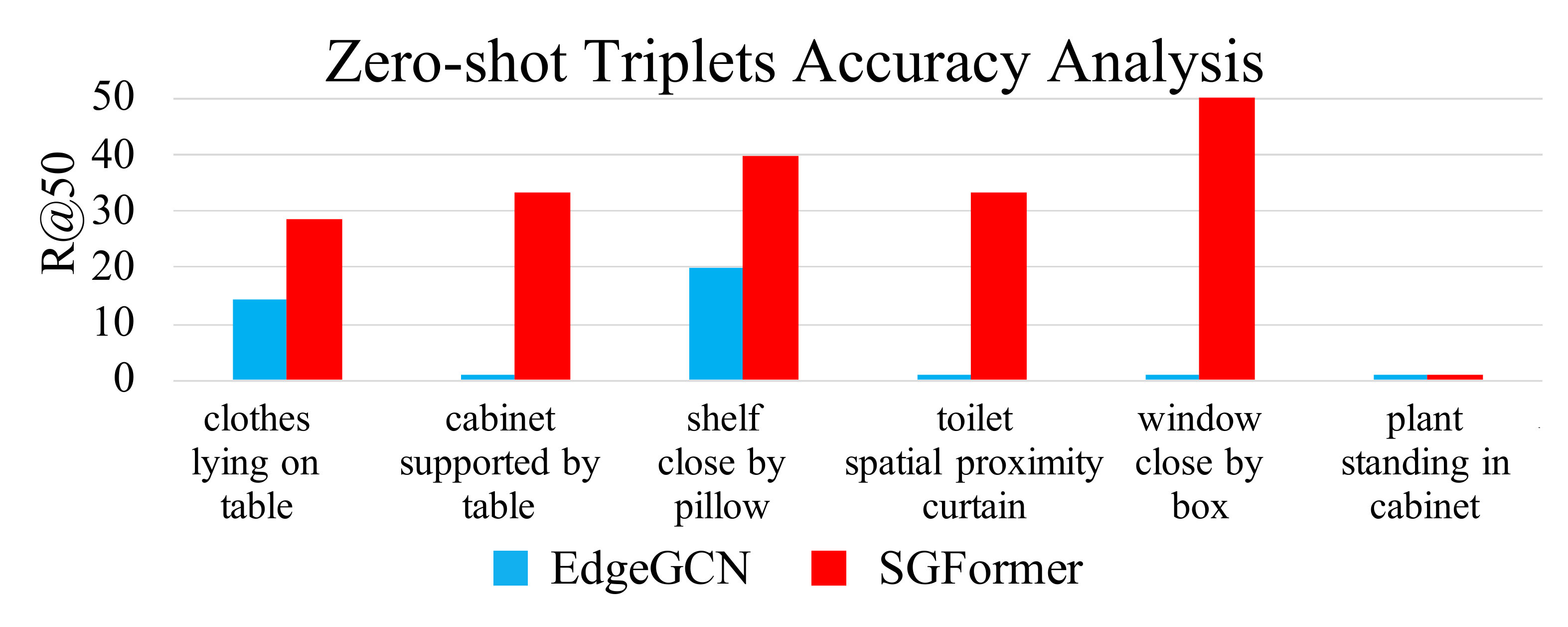}
    \vspace{-6mm}
    \caption{The R@50 results of  EdgeGCN and our SGFormer in terms of the zero-shot relationships}
    \label{zeroshot}
    \vspace{-4mm}
\end{figure} 
\section{How does SGFormer help?}

\noindent\textbf{Scene size analysis.}
Defining scene complexity as the summation of the number of nodes and relationships in the whole scene, we compare the performance of our approach with EdgeGCN in varying scene scales. As shown in Figure~\ref{Relationship_acc_analysis}, our SGFormer exhibits a growing relative gain along with the heightened scene complexity.

\noindent\textbf{Long-tail issue.}~The long-tail phenomenon is very common yet challenging in scene graph generation, where the model usually misleads an uncommon or rare category in the dataset as a common label. As shown in Figure~\ref{fig: long-tail-analysis}, the distribution of each label in the train set is very unbalanced.
Our proposed SGFormer demonstrates superior performance compared to the model without semantic knowledge injection from Table~\ref{table:SIM}. Especially, $ \text{SIL}_{\text{GPT w/ template}}$ achieves the best result, which improves 10.21\% and 22.13\% than baseline w.r.t Node mR@5 and Edge mR@3, indicating that our proposed SIL can exploit the emergent ability of LLM to improve the predictions performance in marginally sampled object and predicate categories, and successfully alleviate the long-tail challenge.  Figure~\ref{fig: long-tail-analysis} shows the improvement is increasingly greater for very ``tail'' items. The analysis of predicates can yield the same conclusions.

\noindent\textbf{Zero-shot scenario.}~
In Figure~\ref{zeroshot}. We can observe that our proposed model can improve by a large margin compared with EdgeGCN~\cite{edge-gcn}, such as $<$cabinet-supported by-table$>$ and $<$window-close by-box$>$. These findings consistently demonstrate the efficacy and superiority of the proposed GEL and SIL in our SGFormer. However, both methods perform poorly in $<$plant-standing in-cabinet$>$ due to this relationship being uncommon in real life, and more training data can resolve this case.

\section{Conclusion}

In this paper, we proposed a novel Semantic Graph Transformer model (\textit{i.e.}, SGFormer) for 3D scene graph generation, including Graph Embedding Layer to capture the global-level structure and Semantic Injection Layer to enhance the objects' visual features with LLMs' powerful semantic knowledge. Extensive experiments on the 3DSSG benchmark demonstrate the effectiveness and state-of-the-art performance of our proposed model. Notably, our SGFormer performs exceptionally well on complicated scenes, and under challenging long-tail and zero-shot scenarios.

\section{Acknowledgement}
This work was partly supported by the Funds for Creative Research Groups of China under Grant 61921003, the National Natural Science Foundation of China under Grant 62202063, the Young Elite Scientists Sponsorship Program by China Association for Science and Technology (CAST) under Grant 2021QNRC001, the 111 Project under Grant B18008, the Open Project Program of State Key Laboratory of Virtual Reality Technology and Systems, Beihang University (No.VRLAB2022C01).
\bibliography{aaai24}

\end{document}